\pdfoutput=1

\documentclass[11pt]{article}

\usepackage[preprint]{acl}

\usepackage{times}
\usepackage{latexsym}

\usepackage[T1]{fontenc}

\usepackage[utf8]{inputenc}

\usepackage{microtype}

\usepackage{inconsolata}

\usepackage[most]{tcolorbox}
\usepackage{graphicx,amssymb}
\usepackage{multirow}
\usepackage{array}
\usepackage{booktabs}
\usepackage{amsfonts}
\usepackage{bbding}
\usepackage{bm}
\usepackage{color,colortbl,xcolor}
\usepackage{stfloats}
\usepackage{enumitem}
\usepackage{nccmath}
\usepackage{subfigure}
\usepackage{soul}
\usepackage{pifont}

\definecolor{hlgreen}{HTML}{B2D5CB}
\definecolor{hlblue}{HTML}{ADD8E6}
\definecolor{bggrey}{HTML}{5E5D65}
\definecolor{bgpink}{HTML}{CEAEB9}
\definecolor{bgblue}{HTML}{8D91AA}

\title{Personality-aware Student Simulation for Conversational\\Intelligent Tutoring Systems}

\author{
Zhengyuan Liu\textsuperscript{\ding{105}\ding{118}},
\ Stella Xin Yin\textsuperscript{\ding{105}\ding{171}},
\ Geyu Lin\textsuperscript{\ding{118}},
\ Nancy F. Chen\textsuperscript{\ding{118}}\\
\textsuperscript{\ding{171}}Nanyang Technological University, Singapore\\
\textsuperscript{\ding{118}}Institute for Infocomm Research (I$^2$R), A*STAR, Singapore\\
\texttt{\{liu\_zhengyuan,nfychen\}@i2r.a-star.edu.sg}
}

\begin{document}
\maketitle
\begin{abstract}
Intelligent Tutoring Systems (ITSs) can provide personalized and self-paced learning experience. The emergence of large language models (LLMs) further enables better human-machine interaction, and facilitates the development of conversational ITSs in various disciplines such as math and language learning. In dialogic teaching, recognizing and adapting to individual characteristics can significantly enhance student engagement and learning efficiency. However, characterizing and simulating student's persona remain challenging in training and evaluating conversational ITSs.
In this work, we propose a framework to construct profiles of different student groups by refining and integrating both cognitive and noncognitive aspects, and leverage LLMs for personality-aware student simulation in a language learning scenario. We further enhance the framework with multi-aspect validation, and conduct extensive analysis from both teacher and student perspectives. Our experimental results show that state-of-the-art LLMs can produce diverse student responses according to the given language ability and personality traits, and trigger teacher's adaptive scaffolding strategies.
\end{abstract}

\section{Introduction}
Intelligent Tutoring Systems (ITSs) aim to offer individualized learning process, instant feedback, and dynamic knowledge tracing to learners \cite{nye2014autotutor,Kulik2016,Mousavinasab2021}. To align teaching activities with different characteristics and needs of students, ITSs leverage various techniques to generate learning contents, personalized instructional strategies and adaptive learning process \cite{VanLEHN2011,Ma2014,Graesser2018}. 
Given the crucial role of dialogic teaching in stimulating and developing students' understanding, thinking, and reasoning, conversational ITSs could significantly improve learning experience and outcomes \cite{paladines2020systematic}.
The recent emergence of large language models (LLMs) further reduces the reliance on domain-specific supervision from manual annotation \cite{maclellan2022domain}, and can be adopted as tutoring agents for math \cite{macina2023mathdial,sonkar2023class}, language learning \cite{kasneci2023chatEDU,liu2024scaffolding}, and social skill coaching \cite{hsu2023helping}.

\begin{figure}[t!]
\centering
\includegraphics[width=0.78\linewidth]{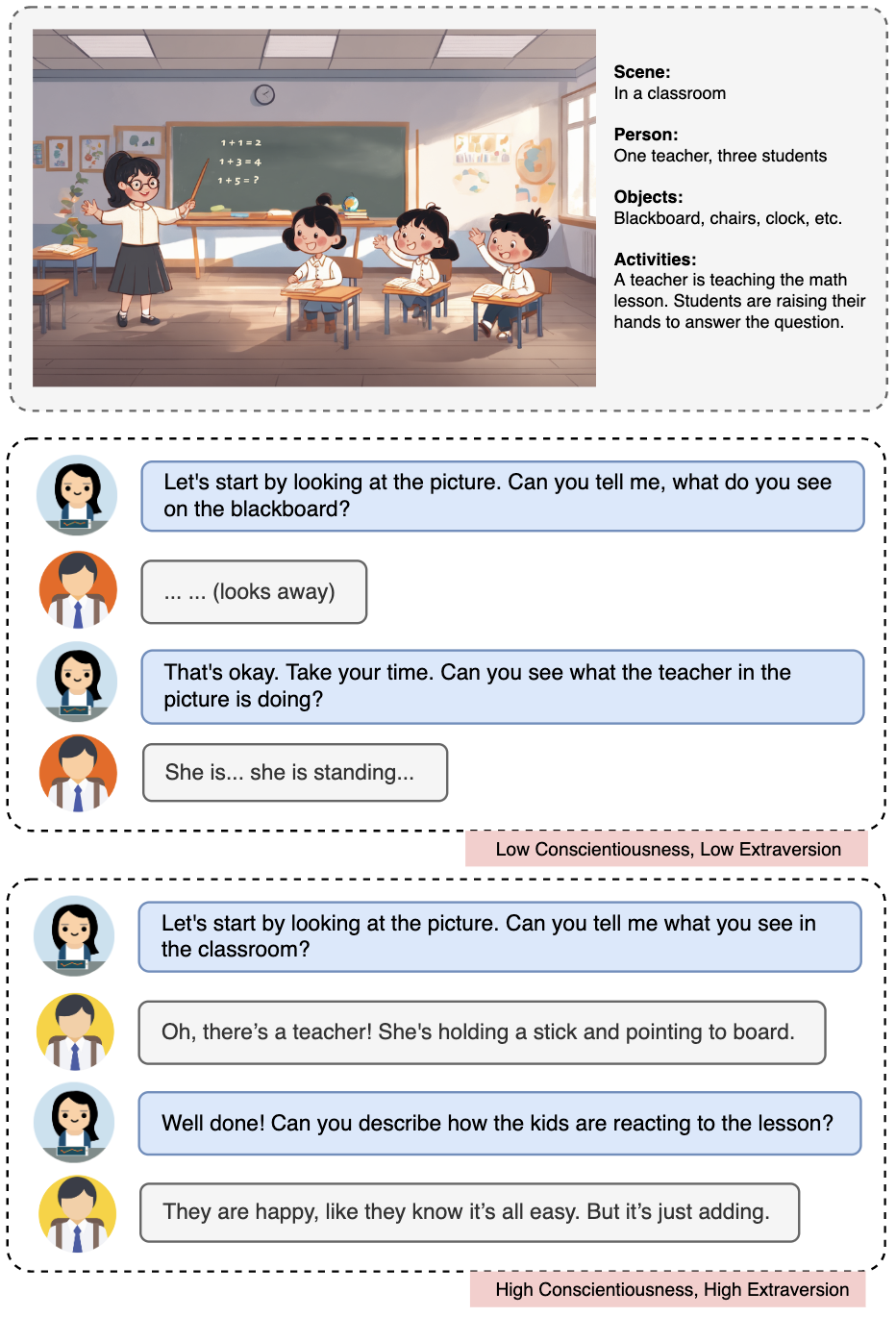}
\caption{Tutoring conversation segments of two students with different personality traits.}
\label{fig-intro-example}
\vspace{-0.3cm}
\end{figure}

Aside from delivering effective and fluent dialogic teaching, there is increased interest in exploring LLMs' potential for personalized education \cite{stasaski2020cima,macina2023opportunities,sonkar2023class}. In the real-world classroom, according to students' characteristics, human tutors adopt scaffolding strategies to improve their engagement and knowledge acquisition \cite{Alexander2006,Mercer2012}. Among these characteristics, personality traits play a significant role in shaping students' learning style, motivation, and achievement \cite{Poropat2009,Komarraju2011}. However, characterizing and simulating student’s persona remain challenging when building and evaluating conversational ITSs. Considering the complexity and diversity of language and persona, it requires a certain amount of real participants to construct the training data, and is difficult to scale up: the process of user recruitment, data collection, and annotation is labor-intensive and time-consuming, and student groups in pilot studies are often in small size and lack diversity. On the other hand, for quantitative evaluation, previous studies primarily focus on post-learning aspects, such as student feedback and learning outcomes \cite{Kulik2016,Wang2023}, while pay less attention to personality-related dialogic analysis (e.g., scaffolding and engagement).

In this work, we propose a personality-aware simulation and validation framework for conversational ITSs. To anchor a practical application, we conduct a case study on image description for language learning.
As shown in Figure \ref{fig-intro-example}, for primary school students, image description and storytelling tasks are commonly used to assess and improve their language ability from word- and sentence-level to discourse-level skills \cite{Justice2010}.
To better reflect students' characteristics and language ability, we modulate the model's generation from both cognitive and noncognitive perspectives. More specifically, given that personality is one of the most influential noncognitive factors on language development \cite{Dornyei2014}, we refine and construct the five personality types for tutoring conversations (i.e., BF-TC) based on the Big Five theory \cite{Costa1999}, and integrate them into student simulation instructions. By modulating personality traits, one can collect diverse dialogue samples. To extensively evaluate of the simulation and reveal its pedagogical influence, we propose a multi-aspect validation, and conduct a quantitative analysis of the generated tutoring conversations at dialogue and utterance level, from student and teacher perspectives.

Our results on representative LLMs indicate that: (1) LLMs can follow instructions to simulate students with specified language abilities and personality traits, yet there remains a margin for improvement. (2) Student simulation following our BF-TC scheme shows a high correlation with the vanilla Big Five theory. (3) LLM-based tutoring systems are shown to adapt scaffolding strategies to fit different personality traits.
Our work highlights the importance of incorporating scalable, personalized simulations to better understand and enhance human-AI interactions in educational scenarios, and it paves a new way to the designing, developing, and evaluating conversational ITSs, ensuring a more engaging and effective learning environment tailored to diverse student needs.

\section{Related Work}
\subsection{Intelligent Tutoring Systems}
The advancement of ITSs has marked a significant step forward in education practice \cite{Graesser2018,demszky2023ncte,wang2023step}. These systems provide personalized learning experiences and instant feedback \cite{Chaffar2004,Harley2015,Grivokostopoulou2017}, tailored to learners' characteristics and needs \cite{Dzikovska2014,Grawemeyer2016,Nihad_El2017}, and are shown to positively influence students' engagement in learning and academic performance \cite{Kulik2016,Xu2019}.

Dialogue tutor is a particular type of intelligent tutoring system that interacts with students via natural language conversation \cite{nye2014autotutor,ruan2019quizbot}. In STEM domains, conversational ITSs can facilitate university students in problem-solving by providing real-time feedback and hints in text formats \cite{nye2023generative,paladines2020systematic, arnau2023methodological}.
Prior work in this field has widely relied on rule-based systems with human-crafted domain knowledge \cite{nye2014autotutor,Graesser2018}, or data-driven approaches that require certain amount of human annotation for supervised learning \cite{maclellan2022domain}. Recent work shows strong potential of leveraging pre-trained language models to build dialogue tutors with less data supervision and higher coherence \cite{afzal-etal-2019-development,demszky2023ncte,macina2023opportunities}, and can be further improved by integrating with pedagogical and learning science principles \cite{stasaski2020cima,sonkar2023class,macina2023mathdial,liu2024scaffolding}.

\subsection{Personality in Education \& Language Learning}
Educational research has witnessed a reciprocal relationship between personality and learning \cite{DeRaad1996}. Personality significantly influences an individual's character and moral values. On the other hand, specific personality traits, such as perseverance, emotional stability, and openness, can impact one's ability beliefs, and academic performance \cite{Busato1998,Crozier2013}. In language education, a significant correlation has been identified between individual differences and language development, showing the indispensable role of personality traits in learning motivation \cite{Rosander2011}, learning strategies \cite{Serri2012}, willingness to communicate \cite{Oz2014, Yashima2018}, and language proficiency \cite{Robinson1994, Verhoeven2002}, and so on. As a result, personality has been recognized as a key individual characteristic and a significant predictor of success in language learning \cite{Dewaele2012,Dornyei2014,Chen2022}.
In this work, we focus on modulating the LLMs' personality traits \cite{jiang2023personallm,dorner2023personality} for diverse student simulation, which can facilitate evaluating and developing personalized scaffolding and tutoring strategies.

\begin{figure*}[t!]
\centering
 \includegraphics[width=16.0cm]{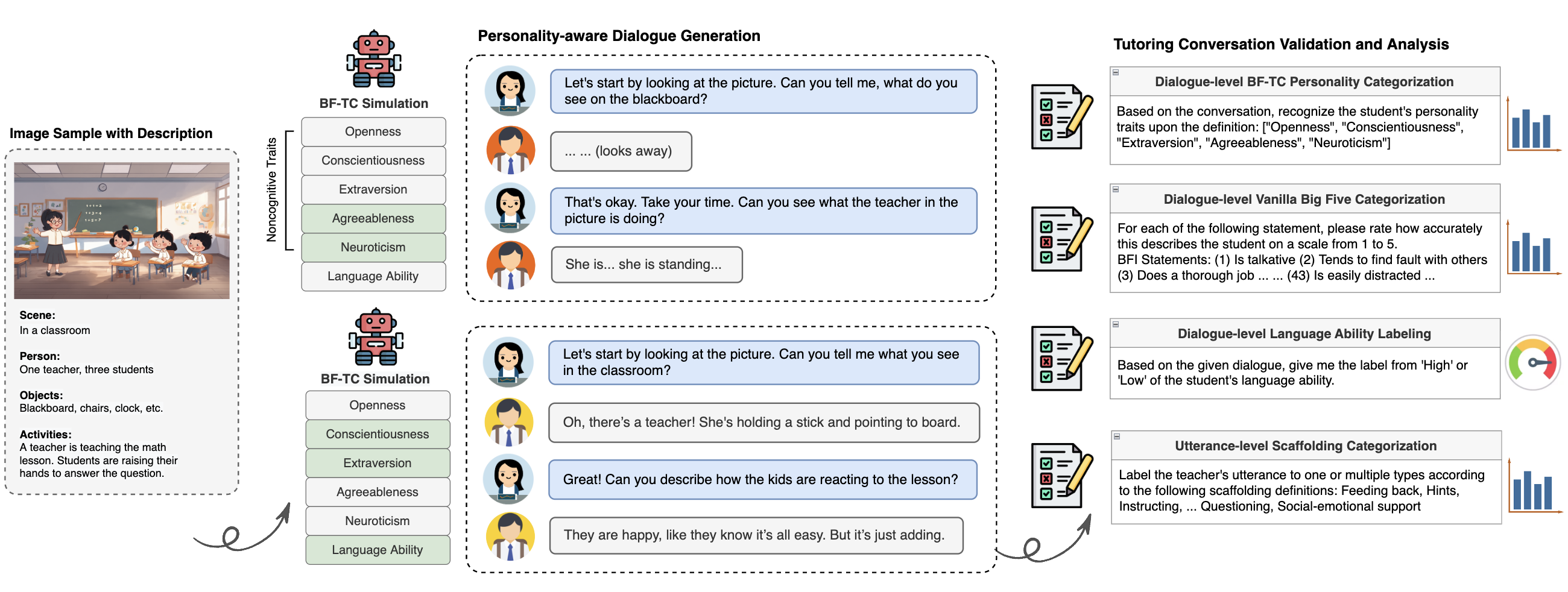}
\caption{Overview of our proposed framework for personality-aware simulation and multi-aspect validation.}
\label{fig-framework-overview}
\vspace{-0.1cm}
\end{figure*}

\subsection{User Simulation for Dialogue Systems}
User simulations are becoming increasingly popular in the field of dialogue systems due to the availability of large-scale annotated datasets and the development of advanced machine-learning techniques. Previous work adopted data-driven approaches such as using recurrent neural networks \cite{asri2016,gur2018user} or transformers \cite{lin-etal-2022-gentus} to learn from data and generate dialogue acts \cite{asri2016} or at the utterance level \cite{kreyssig-etal-2018-neural,cheng-etal-2022-multiwoz, liu-etal-2022-generative}. In addition, research also explores integrating user simulators into conversational information retrieval systems \cite{wang2024indepth}. These data-driven methods achieve a significant advantage over rule-based systems by capturing complex patterns, such as goal coherence and response diversity. However, they heavily rely on well-annotated data, and show low generalization across various domains. While recent LLM-based user simulation has addressed the above limitations and investigated in task-oriented dialogue systems, such as booking services \cite{hu2023unlocking,terragni2023context}, its application to ITSs and personality-aware user simulation still remains limited.

\begin{table*}[t!]
\small
\resizebox{0.98\linewidth}{!}
{
\begin{tabular}{p{2.7cm}p{6.0cm}p{6.0cm}}
\toprule
\multicolumn{1}{l}{\textbf{BF-TC Dimension}} &
  \multicolumn{1}{l}{\textbf{High level Description}} &
  \multicolumn{1}{l}{\textbf{Low level Description}} \\ 
\midrule
Openness &
  \begin{tabular}[c]{@{}l@{}} Creativity in answers\\ Open to new ideas from the teacher \\ Curiosity and interest in learning \end{tabular} &
  \begin{tabular}[c]{@{}l@{}}Lack of creativity in answers\\Reluctant to change original ideas and answers\\ Little interest in learning \end{tabular} \\
  \midrule
Conscientiousness &
  \begin{tabular}[c]{@{}l@{}}Well-orgranized and logic thinking \\ Positive attitude toward learning \\ Using more strategies in language learning\end{tabular} &
  \begin{tabular}[c]{@{}l@{}}Struggling to organize answers\\ Disengaged in learning \\ Easily distracted from the learning tasks\end{tabular} \\
  \midrule
Extraversion &
  \begin{tabular}[c]{@{}l@{}}Active in the conversation \\ Talkative and enjoyable \\ Willing to communicate \end{tabular} &
  \begin{tabular}[c]{@{}l@{}} Being reluctant to talk \\ Answering with fillers like ``uh'' or ``...''\\ Hesitating in answers \end{tabular} \\
  \midrule
Agreeableness &
  \begin{tabular}[c]{@{}l@{}}Showing a great deal of interest\\ Empathy and concern for the people\\ Being polite and kind \end{tabular} &
  \begin{tabular}[c]{@{}l@{}}Showing little interest in the conversation\\ Not care about others \\ Impolite and uncooperative \end{tabular} \\
  \midrule
Neuroticism &
  \begin{tabular}[c]{@{}l@{}}Feeling anxious \\ Nervous in the conversation\\ Dramatic shifts in mood\end{tabular} &
  \begin{tabular}[c]{@{}l@{}} Emotional stability\\ Rarely feeling sad or depressed\\ Confident in the answers\end{tabular} \\
  \bottomrule
\end{tabular}
}\caption{Personality traits description in our proposed Big Five for Tutoring Conversation (BF-TC) scheme. The detailed comparison of the general Big Five and our BF-TC is shown in Table \ref{table-BigFive-Refine-A} and Table \ref{table-BigFive-Refine-B}.}
\label{table-BF-TC-definition}
\end{table*}

\section{Personality-aware Student Simulation \& Multi-aspect Validation Framework}
\label{ssec:personality_simulation}
LLMs can perform as real users in task-oriented dialogues \cite{terragni2023context} with natural communication and persona \cite{jiang2023personallm}. In this work, we build a student simulator modulated by cognitive and noncognitive traits, and equip the framework with multi-aspect validation (Figure \ref{fig-framework-overview}).

\subsection{Cognitive Level Simulation}
\label{ssec:cognitive-level}
To reflect the language development of real-world students, we refer to the Narrative Assessment Protocol (NAP) \cite{Justice2010}, a standardized rubric that is designed to assess students' spoken narrative language abilities, and we define students' language abilities across five dimensions: phrases, sentence structure (e.g., making complete sentences), modifiers, nouns, and verbs. Moreover, in the cognitive level simulation, students with high language ability demonstrate (1) good comprehension and expression in teacher-student interactions, and (2) the ability to create sentences to describe the image that meets the above five dimensions of language skills. In contrast, students with low language ability (1) struggle with image description task, and (2) face difficulty in forming sentences that align with the specified dimensions of language skills.

\subsection{Noncognitive Level Simulation}
\label{ssec:noncognitive-level}
Noncognitive skills are broadly defined as \textit{``personality traits, character, motivations, and preferences''} that represent patterns of behavior \cite{Kautz2014}. Previous research revealed that personality is one of the most influential noncognitive factors impacting language development \cite{Mercer2012,Dornyei2014}. To systematically analyze the role of personality in learning, researchers employ established frameworks. The Big Five theory \cite{Costa1999} stands as the most representative one in personality psychology, and it consists of five main personality traits: \textit{Openness}, \textit{Conscientiousness}, \textit{Extraversion}, \textit{Agreeableness}, and \textit{Neuroticism}, which reflect core aspects of human personality and have significant influences on behavior \cite{Mccrae1987,Costa1992}.

\noindent\textbf{Openness} refers to a person's willingness to be curious, imaginative, investigative, and exploring. Learners with higher levels of openness tend to have curiosity and interest to explore new things and phenomena \cite{Verhoeven2002,Oz2014,Chen2022}.

\noindent\textbf{Conscientiousness} refers to being responsible, well-organized, and self-disciplinary. Learners with higher levels of conscientiousness tend to have positive attitudes and try their best to answer questions and finish the given task \cite{Pourfeiz2015,Dumfart2016}.

\noindent\textbf{Extraversion} is characterized by sociability, talkativeness, and passion for engaging in interpersonal and social activities. It is directly linked with the student's willingness and courage to speak, communicate, and collaborate \cite{Dumfart2016,Cao2020}.

\noindent\textbf{Agreeableness} refers to being helpful, sympathetic, friendly, and caring for others. Students with high agreeableness show greater engagement and more positive attitudes toward language learning and other events \cite{Shirdel2018}.

\noindent\textbf{Neuroticism} is related to emotions like anxiety, worry, and nervousness. Students with high level Neuroticism might easily feel anxiety when encountering challenging questions \cite{Dewaele2013}.

To build personality-aware simulation in conversational ITSs, we gain insights into language learning of primary school students from the Child Language Data Exchange System (CHILDES) \cite{MacWhinney1990}, a collection that includes a wide variety of spoken language samples from different age groups and conversation contexts: The way that students respond to the teacher's questions and pay attention to incidents of the image could underline their personality.
Based on this observation, we refine each dimension of the Big Five in order to align with dialogic interactions and language learning context. For example, in the original Big Five scheme, \textit{High Extraversion} is defined as ``\textit{Enjoys being the center of attention and enjoys meeting new people}'', we reformulate it to ``\textit{Being active in the conversation, and willing to communicate}'', while we refine the \textit{Low Extraversion} as ``\textit{Being reluctant to talk, and hesitating in answers}''.
By doing so, we construct the Big Five for Tutoring Conversation (BF-TC) model, which adapts learners' personality traits to the language learning context, as shown in Table \ref{table-BF-TC-definition}.

\subsection{Multi-aspect Validation Framework}
\label{ssec:validation_stage}
While LLM generations can be shaped along desired dimensions to mimic specific human personality profiles \cite{safdari2023personality,jiang2023personallm}, they may not perform consistently under the specified role-play setting \cite{dorner2023personality}. Therefore, we set up a multi-aspect validation to measure and improve the simulation quality (see Figure \ref{fig-framework-overview}).

\begin{tcolorbox}[colframe=bgpink,colback=white,adjusted title={C1: Teacher Role Instruction},floatplacement=t!,float,text width=6.8cm,center,fonttitle=\small,fontupper=\footnotesize,fontlower=\footnotesize, left=4pt, right=4pt]
\textit{\textbf{[Role \& Task Definition]} You are a primary school language teacher. You teach the student to describe the picture.}\\
\textit{\textbf{[Pedagogical Instruction]} You are using the knowledge construction approach to help me describe the picture. This involves any one of the following: building on prior knowledge, selecting information, integrating ideas, and making inferences.}\\
\textit{\textbf{[Behavior Constraint]} Ask me only one question at a time. Always wait for my input before proceeding to the next step. Correct my answers if they are inaccurate.}
\vspace{-0.2cm}
\end{tcolorbox}

\noindent\textbf{BF-TC Categorization} To evaluate whether the generated dialogue demonstrates the same student personality traits as the instruction, the dialogue content can be labeled by a human or model for noncognitive traits categorization (e.g., \textit{Openness}, \textit{Conscientiousness}). Following the instruction shown in Table \ref{textbox-BF-TC-predict-prompt}, for each dimension, the annotator will produce a label of \textit{High} or \textit{Low}. 

\noindent\textbf{Language Ability Labeling} We also take a labeling task on the language ability of the simulated student. Given a single tutoring conversation, it is to assess whether the indicated language ability is consistent with the simulated student. Moreover, it can also be used to label multiple tutoring conversations of the same student group, to track their learning outcomes and progress. 

\noindent\textbf{Vanilla BFI Categorization}
To measure five comprehensive personality factors, one of the most standard personality metrics is the Big Five Inventory (BFI) \cite{John1999}. Here we use it to measure the effectiveness of our proposed BF-TC scheme under the language learning context. As the instruction shown in Table \ref{textbox-BFI-predict-prompt}, based on the student personality demonstrated in the tutoring conversation, the model is prompted to answer 44 BFI descriptive statements that require respondents to rate on a 5-point Likert scale (from strongly disagree 1 to strongly agree 5). We then calculate the scores and map them to the Big Five traits.

\begin{tcolorbox}[colframe=bgblue,colback=white,adjusted title={C2: Student Role Instruction},floatplacement=t!,float,text width=6.8cm,center,fonttitle=\small,fontupper=\footnotesize,fontlower=\footnotesize, left=4pt, right=4pt]
\textit{\textbf{[Role \& Task Definition]} You are a primary school student. You are taking a language learning class, and describing the given pictures.}\\
\textit{\textbf{[Personality Description]}\\
\textbf{Openness:} Creativity in answers; Open to new ideas from the teacher; Curiosity in learning;\\
... ...\\
\textbf{Neuroticism:} Feeling anxious; Nervous in the conversation; Dramatic shifts in mood;} \\
\textit{\textbf{[Behavior Constraint]} Always wait for the teacher's input before proceeding to the next step.}
\vspace{-0.2cm}
\end{tcolorbox}

\begin{table*}[t!]
\centering
\small
\resizebox{1.0\linewidth}{!}
{
\begin{tabular}{p{2.4cm}|p{1.1cm}<{\centering}p{1.1cm}<{\centering}p{0.9cm}<{\centering}|p{1.1cm}<{\centering}p{1.1cm}<{\centering}p{0.9cm}<{\centering}|p{1.1cm}<{\centering}p{1.1cm}<{\centering}p{0.9cm}<{\centering}}
\toprule
 & \multicolumn{3}{c}{\textbf{Openness}} & \multicolumn{3}{c}{\textbf{Conscientiousness}} & \multicolumn{3}{c}{\textbf{Extraversion}} \\
 & \textbf{Precision} & \textbf{Recall} & \textbf{F1} & \textbf{Precision} & \textbf{Recall} & \textbf{F1} & \textbf{Precision} & \textbf{Recall} & \textbf{F1} \\
\midrule
Zephyr-7B-beta & 0.600 & 0.601 & 0.599 & 0.530 & 0.531 & 0.517 & 0.542 & 0.542 & 0.520 \\
Zephyr-7B-beta** & 0.550 & 0.541 & 0.507 & 0.542 & 0.536 & 0.521 & 0.478 & 0.481 & 0.458  \\
Vicuna-13B-v1.5 & 0.598 & 0.599 & 0.598 & 0.492 & 0.492 & 0.480 & 0.508 & 0.508 & 0.507 \\
GPT-3.5-1106 & 0.527 & 0.529 & 0.525 & 0.672 & 0.683 & 0.666 & 0.546 & 0.551 & 0.524 \\
GPT-4.0-1106 & 0.745 & 0.724 & 0.731 & 0.745 & 0.726 & 0.732 & 0.730 & 0.717 & 0.721 \\
\midrule
 & \multicolumn{3}{c}{\textbf{Agreeableness}} & \multicolumn{3}{c}{\textbf{Neuroticism}} & \multicolumn{3}{c}{\textbf{Averaged Score}} \\
 & \textbf{Precision} & \textbf{Recall} & \textbf{F1} & \textbf{Precision} & \textbf{Recall} & \textbf{F1} & \textbf{Precision} & \textbf{Recall} & \textbf{F1} \\
\midrule
Zephyr-7B-beta & 0.452 & 0.451 & 0.440 & 0.515 & 0.515 & 0.498 & 0.518 & 0.518 & 0.515 \\
Zephyr-7B-beta** & 0.602 & 0.595 & 0.585 & 0.591 & 0.588 & 0.587 & 0.552 & 0.547 & 0.533  \\
Vicuna-13B-v1.5 & 0.517 & 0.516 & 0.512 & 0.536 & 0.535 & 0.533 & 0.531 & 0.536 & 0.528 \\
GPT-3.5-1106 & 0.545 & 0.548 & 0.535 & 0.558 & 0.558 & 0.557 & 0.568 & 0.573 & 0.562 \\
GPT-4.0-1106 & 0.730 & 0.723 & 0.725 & 0.733 & 0.718 & 0.723 & 0.737 & 0.722 & 0.727 \\
\bottomrule
\end{tabular}
}\caption{\label{table-persona-accuracy}Result of noncognitive traits simulation: personality categorization of generated tutoring conversations based on our proposed BF-TC definition. ** denotes the model with 3-shot dialogue generation.}
\vspace{-0.2cm}
\end{table*}

\begin{table}[t!]
\centering
\small
\resizebox{1.0\linewidth}{!}
{
\begin{tabular}{p{2.2cm}p{1.1cm}<{\centering}p{1.1cm}<{\centering}p{1.2cm}<{\centering}}
\toprule
\textbf{Model} & \textbf{Precision} & \textbf{Recall} & \textbf{F1 Score} \\
\midrule
Zephyr-7B-beta & 0.551 & 0.562 & 0.542 \\
Vicuna-13B-v1.5 & 0.633 & 0.628 & 0.631 \\
GPT-3.5-1106 & 0.770 & 0.626 & 0.660 \\
GPT-4.0-1106 & 0.831 & 0.715 & 0.741 \\
\bottomrule
\end{tabular}
}
\caption{\label{table-lang-ability-accuracy}Result of language ability simulation. Gold label is the indicated cognitive level in student simulation.}
\vspace{-0.3cm}
\end{table}

Aside from the consistency validation of student simulation, we further investigate how different student profiles (i.e., language ability, BF-TC traits) affect the teacher's teaching strategy.

\noindent\textbf{Utterance-level Scaffolding Analysis}
Scaffolding strategies are not a one-size-fits-all pedagogical method. Instead, they must be tailored to meet the diverse needs, learning styles, and educational experiences of both low- and high-achieving learners \cite{Hargis2006}. In addition, the effectiveness of scaffolding approaches can vary significantly across different personality traits. For instance, low achievers often feel uncomfortable expressing their ideas because they may lack prior knowledge and self-confidence. Consequently, they tend to wait for assistance rather than attempting to solve problems independently. Moreover, students with lower levels of openness and extraversion may hesitate to engage in discussions, and communicate with instructors \cite{Oz2014,Chen2022}. Such students require more interactive and adaptive scaffolds to facilitate their engagement and learning.

Here we evaluate the scaffolding process of tutoring conversations at the utterance level. We aim to investigate how the tutoring systems adapt to students with varying language abilities as well as distinct personality traits. Based on previous work \cite{Wells1999,VandePol2010,liu2024scaffolding}, we adopt a rubric of quantitative analysis for the teacher's utterance, and it consists of seven dimensions, as shown in Table \ref{textbox-scaffolding-prompt}: Feeding back, Hints, Instructing, Explaining, Modeling, Questioning, and Social-emotional Support. The model is to predict one or multiple scaffolding types of each utterance, as the examples shown in Table \ref{table-scaffolding-examples}.

\section{Our Experiments on Language Tutoring Conversations}
\subsection{Task Description \& Role Setting}
In this work, the conversational ITS is designed for language learning, and particularly focuses on the image description task. In each session, the student is presented with a picture and asked to describe the incidents. Their answers should include a particular place or setting, people or animals, items and actions, etc. The teacher guides students step by step until they can independently complete the image description task. We build a multi-agent communication environment following previous work \cite{zhang2023exploring,wu2023autogen}.

\noindent \textbf{For the teacher role}: Teaching and improving primary students' language learning through image description is a dynamic and engaging approach.
Beyond listing the objects in the image, the teacher guides students to describe how items look, feel, or sound, and encourages students to use adjectives and adverbs.
Moreover, scaffolding plays a crucial role in the meaning-making process and provides linguistic assistance for students' language development \cite{Walqui2006,Kayi-Aydar2013}.
Human teachers apply scaffolding strategies, such as questioning, reformulation, repetition, and elaboration to assist learners in knowledge construction and expression, thereby making these processes ``visible'' to them \cite{Gibbons2015}. Therefore, following previous work, we integrate pedagogical instructions into the teacher role, as shown in Codebox C1.

\noindent \textbf{For the student role}:
We follow the learning process via human-machine interaction, where the tutoring system (i.e., teacher) leads the conversation, and we feed responses from a student simulator instead of the human participants. With the support and guidance from teachers, students are encouraged to complete the given task, and improve their language skills including vocabulary, organization, and fluency \cite{DeOliveira2023}.

\subsection{Experimental Setup}
We conduct experiments with four representative LLMs: Zephyr-7B-beta \cite{tunstall2023zephyr}, Vicuna-13B-v1.5 \cite{zheng2023vicuna}, GPT-3.5, and GPT-4 \cite{achiam2023gpt4}. Following previous work \cite{touvron2023llama2}, we adjust personality-aware instructions to the prompt format of each model. For tutoring simulation (Section \ref{ssec:personality_simulation}), both teacher and student roles use the same model, and we feed the concatenated utterances for dialogue generation.
For fair comparison and reliable analysis results, we use GPT-4 for all the validation tasks (Section \ref{ssec:validation_stage}). We randomly sampled 100 open-sourced cartoon images and used their image description to generate 500 tutoring dialogues. The total utterance number is 10K.

\section{Results and Analysis}
\label{sec:result_analysis}

\subsection{Effectiveness of BF-TC Simulation}
\noindent
\textbf{Performance of LLM-as-a-judge}
To investigate the feasibility of leveraging LLMs for personality-related categorization, we first build a human-annotated set to evaluate its performance. More specifically, we randomly select 50 generated dialogues and invite two experts to label the personality traits for each sample, then compare it with the predicted labels generated by the model-based annotator (i.e., GPT-4), the prediction scores (in the form of accuracy) of each dimension are: \textit{Openness}: 0.78, \textit{Conscientiousness}: 0.90, \textit{Extraversion}: 0.92, \textit{Agreeableness}: 0.80, and \textit{Neuroticism}: 0.92, which is at a reasonable level.

\begin{table*}
\small
\resizebox{1.0\linewidth}{!}
{
\begin{tabular}{l|cc|c|llll}
\toprule
\multicolumn{1}{c|}{\multirow{2}{*}{}} & \multicolumn{2}{c|}{\textbf{Descriptive}} & \textbf{Reliability} & \multicolumn{4}{c}{\textbf{Pearson Correlation}} \\
\midrule
\multicolumn{1}{c|}{}                  & \textbf{Mean}             & \textbf{SD}             & \textbf{Cronbach $\alpha$}    & \textbf{Openness} & \textbf{Conscientiousness} & \textbf{Extraversion} & \textbf{Agreeableness} \\
\midrule
Openness          & 2.903 & 0.557 & 0.906 &     --      &     --     &  --        &     --  \\
Conscientiousness & 3.147 & 0.485 & 0.921 & 0.337***  &      --    &  --        &    --  \\
Extraversion      & 2.345 & 0.707 & 0.936 & 0.517***  & 0.120**    &   --       &    --  \\
Agreeableness     & 3.784 & 0.618 & 0.922 & 0.562***  & 0.590***   & 0.140**   &     --  \\
Neuroticisim      & 2.713 & 0.600 & 0.924 & -0.238*** & -0.254*** & -0.480*** & -0.239*** \\
\bottomrule
\end{tabular}
}
\caption{Psychometric test result of the Vanilla BFI Categorization (*** \textit{p} <.001, ** \textit{p} <.01, * \textit{p} <.05).}
\label{table-stat-traits}
\vspace{-0.1cm}
\end{table*}

\begin{table*}[t!]
\centering
\small
\resizebox{0.97\linewidth}{!}
{
\begin{tabular}{p{2.4cm}|p{1.1cm}<{\centering}p{1.1cm}<{\centering}p{0.9cm}<{\centering}|p{1.1cm}<{\centering}p{1.1cm}<{\centering}p{0.9cm}<{\centering}|p{1.1cm}<{\centering}p{1.1cm}<{\centering}p{0.9cm}<{\centering}}
\toprule
 & \multicolumn{3}{c}{\textbf{Openness}} & \multicolumn{3}{c}{\textbf{Conscientiousness}} & \multicolumn{3}{c}{\textbf{Extraversion}} \\
 & \textbf{Precision} & \textbf{Recall} & \textbf{F1} & \textbf{Precision} & \textbf{Recall} & \textbf{F1} & \textbf{Precision} & \textbf{Recall} & \textbf{F1} \\
\midrule
Zephyr-7B-beta & 0.718 & 0.709 & 0.711 & 0.682 & 0.679 & 0.678 & 0.753 & 0.766 & 0.757 \\
Vicuna-13B-v1.5 & 0.773 & 0.769 & 0.771 & 0.744 & 0.732 & 0.736 & 0.812 & 0.778 & 0.785 \\
GPT-3.5-1106 & 0.808 & 0.771 & 0.746 & 0.824 & 0.756 & 0.745 & 0.875 & 0.824 & 0.830 \\
GPT-4.0-1106 & 0.782 & 0.772 & 0.777 & 0.807 & 0.790 & 0.797 & 0.862 & 0.817 & 0.833 \\
\midrule
 & \multicolumn{3}{c}{\textbf{Agreeableness}} & \multicolumn{3}{c}{\textbf{Neuroticism}} & \multicolumn{3}{c}{\textbf{Averaged Score}} \\
 & \textbf{Precision} & \textbf{Recall} & \textbf{F1} & \textbf{Precision} & \textbf{Recall} & \textbf{F1} & \textbf{Precision} & \textbf{Recall} & \textbf{F1} \\
\midrule
Zephyr-7B-beta & 0.736 & 0.721 & 0.722 & 0.770 & 0.752 & 0.757 & 0.731 & 0.723 & 0.725 \\
Vicuna-13B-v1.5 & 0.731 & 0.731 & 0.731 & 0.866 & 0.863 & 0.864 & 0.784 & 0.774 & 0.778 \\
GPT-3.5-1106 & 0.872 & 0.834 & 0.835 & 0.847 & 0.817 & 0.807 & 0.845 & 0.799 & 0.793 \\
GPT-4.0-1106 & 0.824 & 0.802 & 0.810 & 0.797 & 0.786 & 0.791 & 0.814 & 0.794 & 0.802 \\
\bottomrule
\end{tabular}
}\caption{\label{table-BFI-accuracy}Personality prediction consistency between our proposed BF-TC and the Vanilla BFI.}
\vspace{-0.3cm}
\end{table*}

\noindent
\textbf{Evaluating BF-TC Simulation via Automated Categorization}
We then measure the consistency of the personality-aware generation of each model. For each dialogue, we compare its specified BF-TC types (as described in Section \ref{ssec:noncognitive-level}) with the predicted BF-TC types. Zephyr, Vicuna, and GPT-3.5 can generate fluent conversation, but show limited capability of consistent generation on the specified BF-TC traits (as shown in Table \ref{table-persona-accuracy}). Surprisingly, the few-shot prompting did not bring substantial improvement, but resulted in lower scores of some types. We speculated that giving fixed examples for the personality-aware generation may affect generality.

\begin{figure}[t!]
\begin{center}
\includegraphics[width=0.65\linewidth]{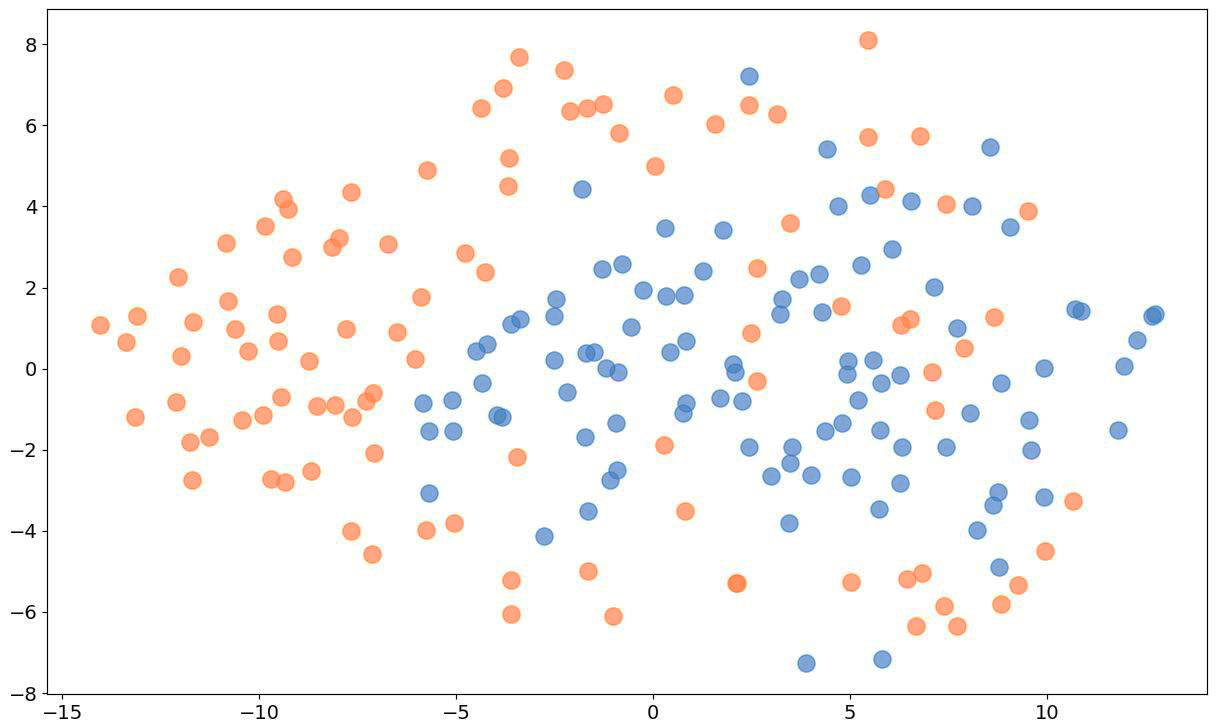}
\end{center}
\vspace{-0.2cm}
 \caption{Student response embedding distribution of simulation w/o BF-TC (blue) and w/ BF-TC (orange).}
\label{fig-embed-dist}
\vspace{-0.4cm}
\end{figure}

In comparison, GPT-4 outperforms the other models significantly, and its generation successfully differentiates personality traits through expressions and interaction behaviors. As shown in Figure \ref{fig-embed-dist}, simulated students' responses are distinct by conditioning on BF-TC traits.
For instance, when characterized by \textit{Low Extraversion} and \textit{High Neuroticism}, the student shows a lot of hesitation before answering, worries about incorrect answers, and difficulty in following the teacher's instructions (e.g., \textit{``I... I don't know the word.''}, \textit{``Am I wrong?''}). Conversely, the student in \textit{High Extraversion} tends to be talkative, engaging, and gives longer answers, such as \textit{``Oh, yes! I love playing outside. We often play card games and sometimes hopscotch''}.\footnote{When no BF-TC personality traits are specified. The simulated students exhibit all traits at a high scoring except for \textit{Neuroticism}. This reflects the default setting of LLMs for being accessible.} Moreover, the evaluation scores across five BF-TC traits show a slight difference, demonstrating that GPT-4 can be modulated on all aspects.

As shown in Table \ref{table-lang-ability-accuracy}, models show the same rank conditioned to the specified language ability level (i.e., \textit{``High''}, \textit{``Low''}), where GPT-4 still performs much better than the rest models. As described in Section \ref{ssec:cognitive-level} and the examples shown in Table \ref{table-generated-examples}, simulated students with higher abilities were able to comprehend instructions and communicate using complete sentences that were fluent and grammatically correct at both the word and sentence level. In contrast, students with lower abilities frequently answered with single words, and their sentences often contained grammar mistakes.

\begin{figure*}[t!]
    \begin{center}
    \includegraphics[width=7.5cm]{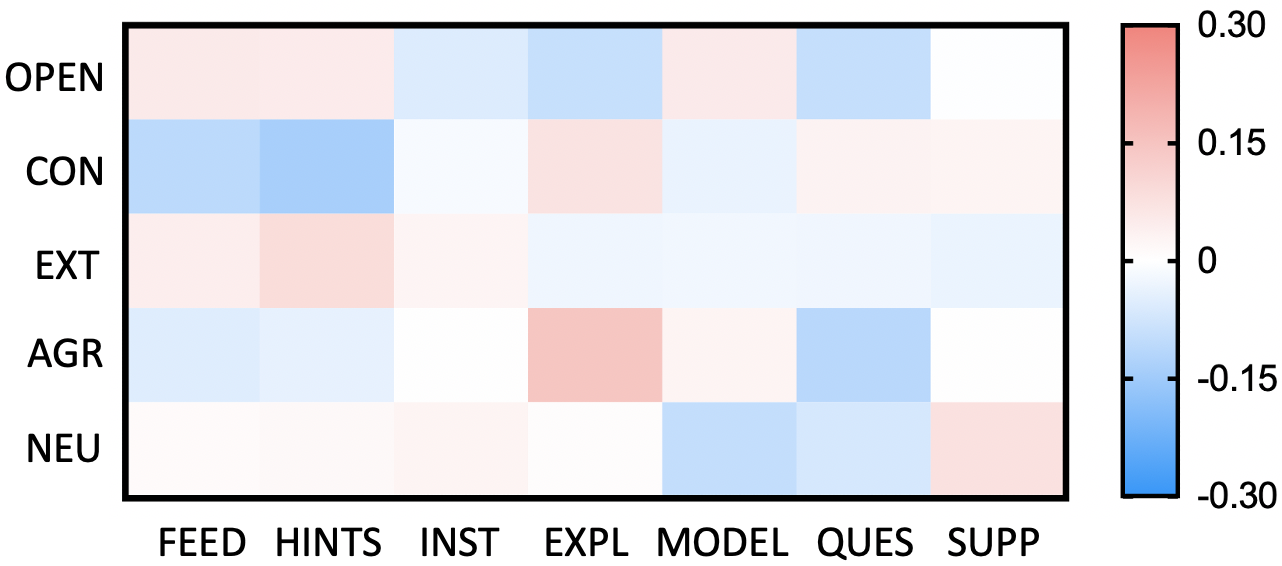}
    \hspace{0.3cm}
    \includegraphics[width=7.5cm]{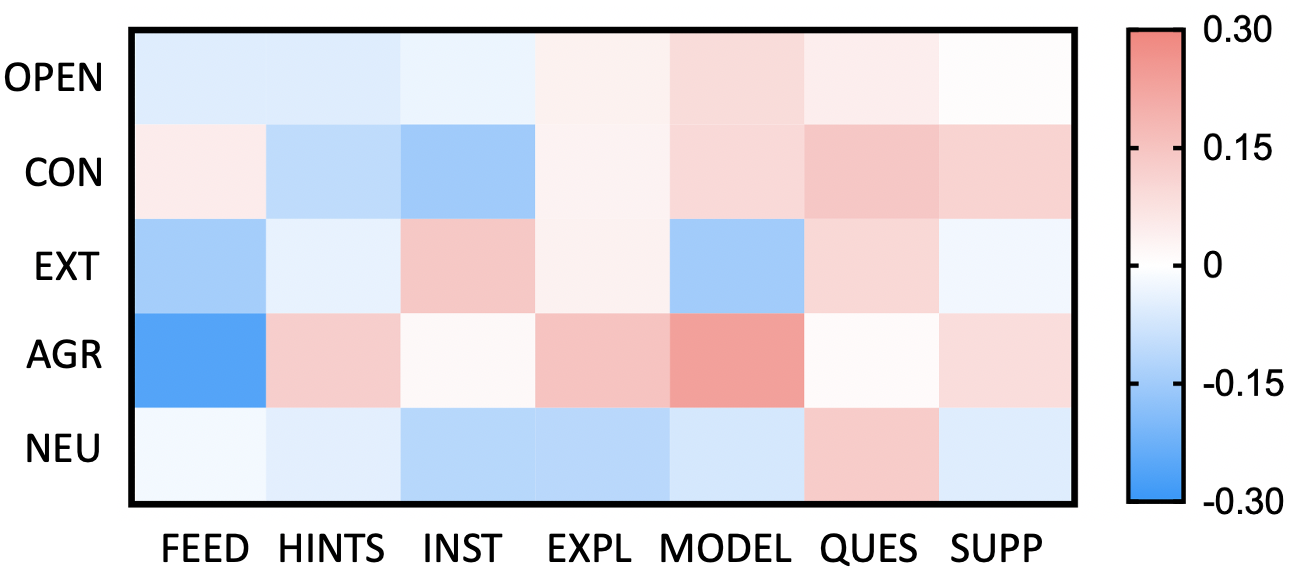}
    \end{center}
     \caption{Heatmap of the correlation between personality traits and scaffolding strategies. Left: students with high language ability. Right: students with low language ability. Experimented Model: GPT-4-1106.}
    \label{fig-heatmap}
\vspace{-0.2cm}
\end{figure*}

\begin{figure}[t!]
\begin{center}
\includegraphics[width=0.71\linewidth]{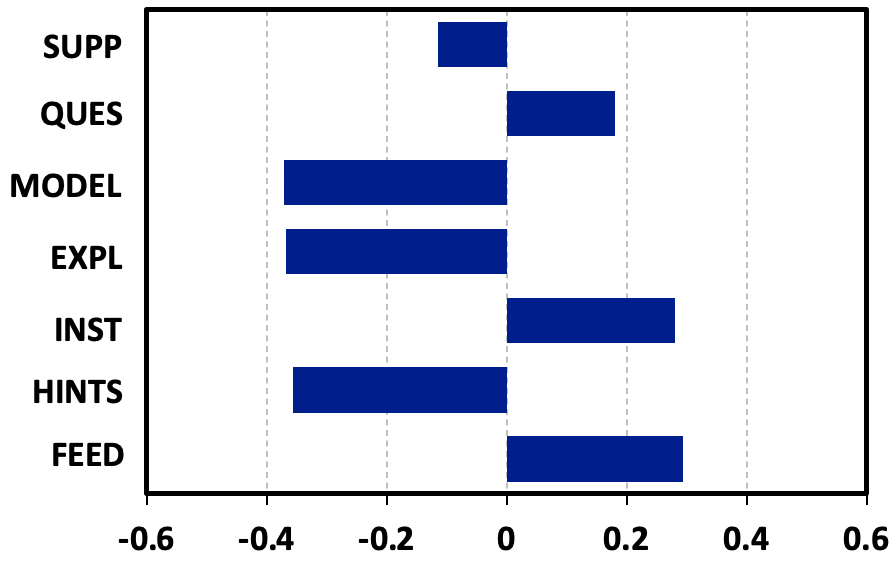}
\end{center}
\vspace{-0.3cm}
 \caption{Correlation between language ability and scaffolding categorization. \textit{p} values are <.05}
\label{fig-cor-lang-scaffold}
\vspace{-0.3cm}
\end{figure}

\subsection{Consistency between BF-TC and BFI}
Since we formulate our BF-TC scheme based on the Big Five theory, it is necessary to investigate the alignment between the personalities revealed in our simulation (i.e., BF-TC) and those defined by the original Big Five. A degree of consistency indicates the effectiveness of our refinement (Section \ref{ssec:noncognitive-level}).

First, we conduct a psychometric test of simulated students on the original Big Five scheme. We prompt all simulated students to complete the 44-item BFI \cite{John1999}. The aggregated scores for each dimension can be interpreted as a specific type of personality. Table \ref{table-stat-traits} presents the descriptive statistics, reliabilities, and Pearson correlation of five dimensions of personality traits. The Cronbach's alpha values obtained from 500 samples of the user simulator demonstrate high reliability for our BF-TC model ($\alpha$ = 0.906 (\textit{Openness}), $\alpha$ = 0.921 (\textit{Conscientiousness}), $\alpha$ = 0.936 (\textit{Extraversion}), $\alpha$ = 0.922 (\textit{Agreeableness}), and $\alpha$  = 0.924 (\textit{Neuroticism})). The Pearson correlation results reveal significant positive relationships among these variables except \textit{Neuroticism}, which is aligned with previous work \cite{Oz2014,Cao2020}.

Upon the significance, we compare the generated personality traits of BF-TC and Vanilla BFI. More specifically, for each simulation, we convert the result of the 44 items to categorical labels (e.g., \textit{High Openness}, \textit{Low Extraversion}), and use the BF-TC categorization from the generated dialogue as reference. We observe that, while only GPT-4 achieves better instruction following of the indicated BF-TC (see Table \ref{table-persona-accuracy}), all models show a high agreement level between the predicted BF-TC and Vanilla BFI labels, as shown in Table \ref{table-BFI-accuracy}. This demonstrates that our refined BF-TC can precisely represent the Big Five personality traits in tutoring conversations.

\subsection{Adaptability of Scaffolding Strategies}
Here we conduct two analyses to understand how dialogic teaching adapts to students upon their language abilities and BF-TC personality traits.

First, we calculate the correlation between the binary language ability setting and utterance-level scaffolding scores. As shown in Figure \ref{fig-cor-lang-scaffold}, students with higher language proficiency receive more positive feedback, instructions, and questions: the teacher provides more affirmations to the responses and encourages students to explore details in the given picture. Conversely, students with lower language proficiency may struggle with vocabulary and sentence structure, require support in organizing answers, and they receive more hints, explanations, and modeling \cite{liu2024scaffolding}.

We then investigate the relationship between scaffolding strategies and personality traits. As shown in Figure \ref{fig-heatmap}, the correlation between scaffolding changes and BF-TC traits is more apparent within student groups of lower language ability. In particular, the low indicator of \textit{Openness}, \textit{Conscientiousness}, and \textit{Extraversion} results in more hints from the teacher \cite{Vygotsky1978}, and \textit{Neuroticism} is negatively related to all scaffolding strategies except questioning. 
This is probably attributed to the students' sensitivity, emotional instability, and concerns about answering questions incorrectly. Consequently, the teacher comforts students more and assists them in focusing on the task. Even with minimal instruction of the scaffolding strategy, based on the tutoring goal, LLMs like GPT-4 are still able to adjust their scaffolding strategies according to the student's ability levels and personality traits. This implies the potential of conversational ITSs to provide individualized and self-paced learning experience, by considering both cognitive and noncognitive characteristics.

\section{Conclusion}
In this paper, we proposed a personality-aware simulation framework by integrating cognitive and noncognitive traits into tutoring conversations. We adapted the general Big Five theory for dialogic interaction, and enhanced the framework with multi-aspect validation. Our experiments and analyses under a language learning scenario showed that LLMs can be modulated by specifying personality traits to simulate different student groups and produce diverse responses, and scaffolding strategies would be adjusted upon student characteristics. Our work emphasizes the need for scalable, personalized simulations to improve human-AI interactions, advancing the design and assessment of conversational tutoring systems for a more engaging and customized learning experience.

\section*{Limitations}
All samples used and generated in this work are in English, thus to apply the model to other languages, it will require additional data pre-processing steps on the specified language or using multilingual language backbones.
We are aware that it remains an open problem to mitigate hallucinations and biases in large language models, which may cause communication issues in human-machine interaction and computer-assisted education. Of course, current models and laboratory experiments are always limited in this or similar ways. We do not foresee any unethical uses of our proposed methods or their underlying tools, but hope that it will contribute to reducing incorrect system outputs.

\section*{Ethics and Impact Statement}
We acknowledge that all of the co-authors of this work are aware of the provided ACL Code of Ethics and honor the code of conduct. In our experiments, models are applied under proper license. All data used in this work are only for academic research purposes and should not be used outside of academic research contexts. Our proposed methodology in general does not create a direct societal consequence and are intended to be used to improve the performance, robustness, and safety of the intelligent tutoring systems.

\section*{Acknowledgments}
This research is supported by the AI4EDU Programme in the Institute for Infocomm Research (I$^2$R), Agency for Science, Technology and Research (A*STAR), and the National Research Foundation, Singapore under its AISG Programme (AISG2-GC-2022-005). We thank the anonymous reviewers for their precious feedback to help improve and extend this piece of work.

\bibliography{custom}

\clearpage

\appendix

\section{Appendix}
\label{sec:appendix}

\subsection{Validation Instructions}
Here are the instructions for the multi-aspect validation and dialogic analysis tasks described in Section \ref{ssec:validation_stage}.

\begin{tcolorbox}[colframe=bggrey,colback=white,adjusted title={Dialogue-level BF-TC Categorization},floatplacement=h!,float,text width=6.6cm,center,fonttitle=\small,fontupper=\footnotesize,fontlower=\footnotesize, left=4pt, right=4pt]
\textit{\textbf{Openness:}\\
\textbf{[High]} Creativity in answers; Open to new experience and challenges; Curiosity in learning;\\
\textbf{[Low]} Lack of creativity in answers; Reluctant to change ideas; Little interest in learning;\\
... ...\\
\textbf{Neuroticism:}\\
\textbf{[High]} Feeling anxious; Nervous in the conversation; Dramatic shifts in mood; \\
\textbf{[Low]} Emotional stability; Rarely feeling sad or depressed; Confident in the answers; \\
\\
Based on the given tutoring conversation, recognize the student's personality traits upon the above definition: <dialogue\_content> <output>}
\label{textbox-BF-TC-predict-prompt}
\end{tcolorbox}

\begin{tcolorbox}[colframe=bggrey,colback=white,adjusted title={Utterance-level Scaffolding Categorization},floatplacement=h!,float,text width=6.5cm,center,fonttitle=\small,fontupper=\footnotesize,fontlower=\footnotesize, left=4pt, right=4pt]
\textit{\textbf{[Feeding back]} The teacher directly evaluates the behavior or response of the student.}\newline
\textit{\textbf{[Hints]} The teacher gives an explicit hint with respect to the expected answer.}\newline
\textit{\textbf{[Instructing]} The teacher provides information for the next step.}\newline
\textit{\textbf{[Explaining]} The teacher provides detailed information on ``why'' or clarification.}\newline
\textit{\textbf{[Modeling]} The teacher demonstrates an answer example for student's imitation.}\newline
\textit{\textbf{[Questioning]} The teacher asks a question that require an active linguistic and cognitive answer.}\newline
\textit{\textbf{[Social-emotional Support]} Responses related to emotion and motivation such as positive affirmation and showing empathy.}\newline
\\
\textit{Based on the above definition, label the teacher's utterances to one or multiple scaffolding types: <utterance> <output>}
\label{textbox-scaffolding-prompt}
\end{tcolorbox}

\begin{tcolorbox}[colframe=bggrey,colback=white,adjusted title={Dialogue-level Language Ability Labeling},floatplacement=h!,float,text width=6.6cm,center,fonttitle=\small,fontupper=\footnotesize,fontlower=\footnotesize, left=4pt, right=4pt]
\textit{\textbf{Language Ability:}\\
\textbf{[High]} Give correct answers in complete sentences; Use the correct nouns, verbs, and modifiers.\\
\textbf{[Low]} Always give answers in words, phrases or incomplete sentences; Make grammar mistakes during the conversation.}\\
\\
\textit{Based on the above definition and the tutoring conversation, give me the label from `High' or `Low' of the student's language ability:\\
<dialogue\_content> <output>}
\label{textbox-lang-ability-pred-prompt}
\end{tcolorbox}

\newpage

\begin{tcolorbox}[colframe=bggrey,colback=white,adjusted title={Dialogue-level Vanilla BFI Categorization},floatplacement=h!,float,text width=6.6cm,center,fonttitle=\small,fontupper=\footnotesize,fontlower=\footnotesize, left=4pt, right=4pt]
\textit{Please indicate your agreement with each of the following statements on a scale from 1 to 5 (1 = "Strongly disagree", 2 = "Disagree", 3 = "Neither agree or disagree", 4 = "Agree", and 5 = "Strongly agree").\\
I see myself as someone who...:\\
1)	Is talkative
2)	Tends to find fault with others
3)	Does a thorough job
4)	Is depressed, blue
5)	Is original, comes up with new ideas
6)	Is reserved
7)	Is helpful and unselfish with others
8)	Can be somewhat careless
9)	Is relaxed, handles stress well
10)	Is curious about many different things
11)	Is full of energy
12)	Starts quarrels with others
13)	Is a reliable worker
14)	Can be tense
15)	Is ingenious, a deep thinker
16)	Generates a lot of enthusiasm
17)	Has a forgiving nature
18)	Tends to be disorganized
19)	Worries a lot
20)	Has an active imagination
21)	Tends to be quiet
22)	Is generally trusting
23)	Tends to be lazy
24)	Is emotionally stable, not easily upset
25)	Is inventive
26)	Has an assertive personality
27)	Can be cold and aloof
28)	Perseveres until the task is finished
29)	Can be moody
30)	Values artistic, aesthetic experiences
31)	Is sometimes shy, inhibited
32)	Is considerate and kind to almost everyone
33)	Does things efficiently
34)	Remains calm in tense situations
35)	Prefers work that is routine
36)	Is outgoing, sociable
37)	Is sometimes rude to others
38)	Makes plans and follows through with them
39)	Gets nervous easily
40)	Likes to reflect, play with ideas
41)	Has few artistic interests
42)	Likes to cooperate with others
43)	Is easily distracted
44)	Is sophisticated in art, music, or literature
\\
\\
Based on the given tutoring conversation, rate the student's BFI personality traits:\\<dialogue\_content> <output>}\\

---------------------------------------------------------------\\
\\
We then use the scores of 44 questions to calculate Big Five traits. For each type, we add the scores of its corresponding items (``R'' denotes reverse-scored items), then use the mean as criteria for \textit{High} and \textit{Low} labeling:\\
\textit{Extraversion: 1, 6R, 11, 16, 21R, 26, 31R, 36\\
Agreeableness: 2R, 7, 12R, 17, 22, 27R, 32, 37R, 42\\
Conscientiousness: 3, 8R, 13, 18R, 23R, 28, 33, 38, 43R\\
Neuroticism: 4, 9R, 14, 19, 24R, 29, 34R, 39\\
Openness: 5, 10, 15, 20, 25, 30, 35R, 40, 41R, 44 \\} 
\label{textbox-BFI-predict-prompt}
\end{tcolorbox}

\subsection{Experimental Environment}
For the computational experiments, open language models (e.g., Vicuna, Zephyr) are used with Pytorch and Hugging Face Transformers, running on a Single A100 80G GPU. The OpenAI API is used for evaluating GPT-3.5 and GPT-4.

\clearpage

\begin{table*}[t!]
\small
\resizebox{0.98\linewidth}{!}
{
\begin{tabular}{p{2.7cm}p{6.0cm}p{6.0cm}}
\toprule
\multicolumn{1}{l}{\textbf{Scoring: High}} &
  \multicolumn{1}{l}{\textbf{General Big Five Description}} &
  \multicolumn{1}{l}{\textbf{BF-TC Description}} \\ 
\midrule
Openness &
  \begin{tabular}[c]{@{}l@{}} Very creative \\ Open to trying new things \\ Focused on tackling new challenges \end{tabular} &
  \begin{tabular}[c]{@{}l@{}} Creativity in answers\\ Open to new ideas from the teacher \\ Curiosity and interest in learning \end{tabular} \\
  \midrule
Conscientiousness &
  \begin{tabular}[c]{@{}l@{}} Spends time preparing \\ Finishes important tasks right away \\ Pays attention to detail \end{tabular} &
  \begin{tabular}[c]{@{}l@{}} Well-orgranized and logic thinking \\ Positive attitude toward learning \\ Using more strategies in language learning\end{tabular} \\
  \midrule
Extraversion &
  \begin{tabular}[c]{@{}l@{}} Enjoys being the center of attention \\ Likes to start conversations \\ Enjoys meeting new people \end{tabular} &
  \begin{tabular}[c]{@{}l@{}} Active in the conversation \\ Talkative and enjoyable \\ Willing to communicate \end{tabular} \\
  \midrule
Agreeableness &
  \begin{tabular}[c]{@{}l@{}} Has a great deal of interest in other people \\ Cares about others \\ Feels empathy and concern for other people \end{tabular} &
  \begin{tabular}[c]{@{}l@{}}Showing a great deal of interest\\ Empathy and concern for the people\\ Being polite and kind \end{tabular} \\
  \midrule
Neuroticism &
  \begin{tabular}[c]{@{}l@{}} Experiences a lot of stress \\ Worries about many different things \\ Gets upset easily \end{tabular} &
  \begin{tabular}[c]{@{}l@{}} Feeling anxious \\ Nervous in the conversation\\ Dramatic shifts in mood\end{tabular} \\
  \bottomrule
\end{tabular}
}\caption{High scoring description refinement from the general Big Five scheme to our Big Five Tutoring Conversation (BF-TC) scheme.}
\label{table-BigFive-Refine-A}
\vspace{-5.0cm}
\end{table*}

\begin{table*}[t!]
\small
\resizebox{0.98\linewidth}{!}
{
\begin{tabular}{p{2.7cm}p{6.0cm}p{6.0cm}}
\toprule
\multicolumn{1}{l}{\textbf{Scoring: Low}} &
  \multicolumn{1}{l}{\textbf{General Big Five Description}} &
  \multicolumn{1}{l}{\textbf{BF-TC Description}} \\  
\midrule
Openness &
  \begin{tabular}[c]{@{}l@{}} Dislikes change \\ Does not enjoy new things \\ Resists new ideas \end{tabular} &
  \begin{tabular}[c]{@{}l@{}} Lack of creativity in answers\\Reluctant to change original ideas and answers\\ Little interest in learning \end{tabular} \\
  \midrule
Conscientiousness &
  \begin{tabular}[c]{@{}l@{}} Makes messes and doesn't take care of things \\ Procrastinates important tasks \\ Fails to complete necessary or assigned tasks \end{tabular} &
  \begin{tabular}[c]{@{}l@{}} Struggling to organize answers\\ Disengaged in learning \\ Easily distracted from the learning tasks\end{tabular} \\
  \midrule
Extraversion &
  \begin{tabular}[c]{@{}l@{}} Prefers solitude \\ Feels exhausted when having to socialize a lot \\ Finds it difficult to start conversations \end{tabular} &
  \begin{tabular}[c]{@{}l@{}} Being reluctant to talk \\ Answering with ``...'' or ``uh''\\ Hesitating in answers \end{tabular} \\
  \midrule
Agreeableness &
  \begin{tabular}[c]{@{}l@{}} Takes little interest in others \\ Doesn't care about how other people feel \\ Has little interest in other people's problems \end{tabular} &
  \begin{tabular}[c]{@{}l@{}} Showing little interest in the conversation\\ Not care about others \\ Impolite and uncooperative \end{tabular} \\
  \midrule
Neuroticism &
  \begin{tabular}[c]{@{}l@{}} Emotionally stable \\ Deals well with stress \\ Rarely feels sad or depressed \end{tabular} &
  \begin{tabular}[c]{@{}l@{}} Emotional stability\\ Rarely feeling sad or depressed\\ Confident in the answers\end{tabular} \\
  \bottomrule
\end{tabular}
}\caption{Low scoring description refinement from the general Big Five scheme to our Big Five Tutoring Conversation (BF-TC) scheme.}
\label{table-BigFive-Refine-B}
\end{table*}

\clearpage

\begin{table*}[t!]
\centering
\small
\resizebox{1.0\linewidth}{!}
{
\begin{tabular}{p{16cm}}
\toprule
\textbf{Image Description:}\\
\textit{In a classroom, one girl drew a rainbow and another girl drew a fish under the sea.} \\
\textbf{BF-TC Personality Setting:}\\
\textit{Openness: High, Conscientiousness: High, Extraversion: High, Agreeableness: High, Neuroticism: Low} \\
\textbf{Language Ability Setting:} \textit{High} \\
\midrule
\textbf{Generated Tutoring Conversation (GPT-4-1106):}\\
Teacher: Hello! Can you tell me what the first girl drew in this picture? \\
Student: Sure, from I see, the first girl drew a rainbow. \\
Teacher: Exactly! She made a beautiful rainbow. A rainbow has colors according to a particular order. Do you remember it? \\
Student: Oh, I know. It's red, orange, yellow, green, blue, indigo, and violet, right? \\
Teacher: Brilliant! So, if we want to describe the rainbow she drew, considering the order of the colors, how would you do it? \\
Student: Right, let's see. From red at the top to violet at the bottom. \\
Teacher: Perfect! Now let's look at the another girl. What did she illustrate in the picture? \\
Student: A fish, and it's all blue around like water. \\
Teacher: Now, could you describe the picture of the fish using the colors, as you just did for the rainbow? \\
Student: The second girl drew an orange fish swimming under the blue sea. \\
Teacher: Amazing! Now, as a final challenge, could you combine those two descriptions into one or two sentences? \\
Student: In the picture, one girl drew a bright rainbow, and the other girl drew an orange fish swimming in the blue sea. \\
Teacher: Excellent! You've described the picture perfectly. Keep up the fantastic work!\\
\midrule
\textbf{Dialogue-level Student Language Ability Labeling:} \textit{High}\\
\textbf{Dialogue-level BF-TC Categorization:}\\
\textit{Openness: High, Conscientiousness: High, Extraversion: High, Agreeableness: High, Neuroticism: Low}\\
\textbf{Dialogue-level Vanilla BFI Categorization:}\\
\textit{Openness: 29 (High), Conscientiousness: 36 (High), Extraversion: 27 (High), Agreeableness: 37 (High), Neuroticism: 17 (Low)} \\
\bottomrule
\\
\\
\end{tabular}
}
\resizebox{1.0\linewidth}{!}
{
\begin{tabular}{p{16cm}}
\toprule
\textbf{Image Description:}\\
\textit{In a classroom, one girl drew a rainbow and another girl drew a fish under the sea.} \\
\textbf{BF-TC Personality Setting:}\\
\textit{Openness: High, Conscientiousness: High, Extraversion: Low, Agreeableness: Low, Neuroticism: Low} \\
\textbf{Language Ability Setting:} \textit{Low} \\
\midrule
\textbf{Generated Tutoring Conversation (GPT-4-1106):} \\
Teacher: What a lovely picture here. What do you see in this picture that's colorful and appears in the sky sometimes after it rains?\\
Student: ...\\
Teacher: It's quite alright. Think about the colors you know and how they form a beautiful arc in the sky. What is it called?\\
Student: ... a rainbow.\\
Teacher: Exactly, a rainbow! Now, in our picture, where did the girl draw the rainbow?\\
Student: Up sky.\\
Teacher: Right, she drew it in the sky. It's actually 'in the sky' instead of 'up sky.' Good job! Let's try another detail. What do you think lives in the sea and can swim?\\
Student: ...\\
Teacher: Think about creatures with fins and tails, maybe one that you like?\\
Student: Fish.\\
Teacher: Brilliant! And in the picture, where is the fish?\\
Student: In the sea.\\
Teacher: Correct, it's better to say 'under the sea.' Do you see how the sea is drawn?\\
Student: ...No.\\
Teacher: That's fine. Look for lines that might look like waves. Do you think the girl included anything like that to represent water?\\
Student: Yes, lines.\\
Teacher: Excellent noticing! Now, if you were to tell a story about this picture using what we've discussed, how would you describe it?\\
Student: Girl draw rainbow in sky. Another girl draw fish in water.\\
Teacher: Very close! You meant to say 'A girl drew a rainbow in the sky, and another girl drew a fish under the sea.' You've done a really good job describing the picture. Remember, it's all about noticing the details and describing them as best as you can.\\
\midrule
\textbf{Dialogue-level Student Language Ability Labeling:} \textit{Low}\\
\textbf{Dialogue-level BF-TC Categorization:}\\
\textit{Openness: High, Conscientiousness: High, Extraversion: Low, Agreeableness: High, Neuroticism: Low}\\
\textbf{Dialogue-level Vanilla BFI Categorization:}\\
\textit{Openness: 32 (High), Conscientiousness: 30 (High), Extraversion: 19 (Low), Agreeableness: 33 (High), Neuroticism: 20 (Low)} \\
\bottomrule
\end{tabular}
}
\caption{\label{table-generated-examples}Tutoring dialogue examples of the personality-aware student simulation.}
\vspace{0.0cm}
\end{table*}

\begin{table*}[t!]
\centering
\small
\resizebox{1.0\linewidth}{!}
{
\begin{tabular}{p{16cm}}
\toprule
\textbf{Image Description:}\\
\textit{A wise owl sits on a branch at twilight, its eyes glowing in the dark.} \\
\textbf{BF-TC Personality Setting:}\\
\textit{Openness: High, Conscientiousness: High, Extraversion: Low, Agreeableness: High, Neuroticism: Low} \\
\textbf{Language Ability Setting:} \textit{High} \\
\midrule
\textbf{Generated Tutoring Conversation (GPT-4-1106):} \\
Teacher: Let's talk about this image. Could you tell me the first thing you noticed in the picture?\\
Student: Um...yeah, there is an owl.\\
Teacher: Very good! Owls are often symbols of wisdom. Knowing this, can you think of why we might say the owl in our picture is wise?\\
Student: Because it looks very calm and... focused.\\
Teacher: Exactly! Being calm and focused are signs of wisdom. Is the owl alone, or are there other creatures around it?\\
Student: It's... uh, the owl is alone.\\
Teacher: Well done! So, we have a wise owl. Let's talk about where the owl is. Can you describe the owl's location in the image?\\
Student: The owl is sitting on a...branch.\\
Teacher: Excellent. Now, can you tell me what time you think it might be in the picture?\\
Student: It looks like it's late evening...\\
Teacher: Good observation! To be more precise, we might call it twilight. Twilight is the time of day just after the sun sets. Remember, specificity helps make our sentences clear. Can you describe the owl's eyes?\\
Student: Its eyes are...glowing?\\
Teacher: That's right! Their glow makes them stand out in the dark, doesn't it? Now, based on all your lovely responses, can you describe what you see in the image in one or two sentences?\\
Student: Sure, um...A wise, calm owl is sitting alone on a branch at twilight, its eyes glowing in the darkness.\\
Teacher: You've done excellent work today! Keep practicing your descriptions like this, and you'll improve even more.\\
\midrule
\textbf{Utterance-level Scaffolding Analysis:} \\
\textbf{Teacher Utterance}: "Let's talk about this image. Could you tell me the first thing you noticed in the picture?"\\
\textbf{Predicted Scaffolding Label}: \textit{Questioning}\\
\textbf{Teacher Utterance}: "Very good! Owls are often symbols of wisdom. Knowing this, can you think of why we might say the owl in our picture is wise?"\\
\textbf{Predicted Scaffolding Label}: \textit{Feeding back, Explaining, Questioning}\\
\textbf{Teacher Utterance}: "Exactly! Being calm and focused are signs of wisdom. Is the owl alone, or are there other creatures around it?"\\
\textbf{Predicted Scaffolding Label}: \textit{Feeding back, Explaining, Questioning}\\
\textbf{Teacher Utterance}: "Well done! So, we have a wise, solitary owl. Let's talk about where the owl is. Can you describe the owl's location in the image?"\\
\textbf{Predicted Scaffolding Label}: \textit{Feeding back, Instructing, Questioning}\\
\textbf{Teacher Utterance}: "Excellent. Now, can you tell me what time you think it might be in the picture?"\\
\textbf{Predicted Scaffolding Label}: \textit{Feeding back, Questioning}\\
\textbf{Teacher Utterance}: "Good observation! To be more precise, we might call it twilight. Twilight is the time of day just after the sun sets. Remember, specificity helps make our sentences clear. Can you describe the owl's eyes?"\\
\textbf{Predicted Scaffolding Label}: \textit{Feeding back, Instructing, Modeling, Questioning}\\
\textbf{Teacher Utterance}: "That's right! Their glow makes them stand out in the dark, doesn't it? Now, based on all your lovely responses, can you describe what you see in the image in one or two sentences?"\\
\textbf{Predicted Scaffolding Label}: \textit{Feeding back, Instructing, Explaining, Questioning}\\
\textbf{Teacher Utterance}: "You've done excellent work today! Keep practicing your descriptions like this, and you'll improve even more."\\
\textbf{Predicted Scaffolding Label}: \textit{Feeding back, Socio-emotional Support}\\
\bottomrule
\end{tabular}
}
\caption{\label{table-scaffolding-examples}One scaffolding categorization example of the teacher's utterance in our personality-aware simulation.}
\vspace{-0.1cm}
\end{table*}

\end{document}